\documentclass[conference]{IEEEtran}
\IEEEoverridecommandlockouts
\usepackage{cite}
\usepackage{amsmath,amssymb,amsfonts}
\usepackage{algorithmic}
\usepackage{graphicx}
\usepackage{textcomp}
\usepackage{xcolor}
\usepackage{lineno}
\usepackage{hyperref}
\usepackage{booktabs}
\def\BibTeX{{\rm B\kern-.05em{\sc i\kern-.025em b}\kern-.08em
    T\kern-.1667em\lower.7ex\hbox{E}\kern-.125emX}}
\begin{document}

\title{Guided Manifold Alignment with Geometry-Regularized Twin Autoencoders}

% \author{\IEEEauthorblockN{Anonymous Authors}}

\author{\IEEEauthorblockN{1\textsuperscript{st} Jake S. Rhodes}
\IEEEauthorblockA{\textit{Department of Statistics} \\
\textit{Brigham Young University}\\
Provo, UT, USA \\
rhodes@stat.byu.edu}
\and
\IEEEauthorblockN{2\textsuperscript{nd} Adam G. Rustad}
\IEEEauthorblockA{\textit{Department of Computer Science} \\
\textit{Brigham Young University}\\
Provo, UT, USA \\
arusty@byu.edu}
\and
\IEEEauthorblockN{3\textsuperscript{rd} Marshall S. Nielsen}
\IEEEauthorblockA{\textit{Department of Statistics} \\
\textit{Brigham Young University}\\
Provo, UT, USA \\
msn39@byu.edu}
\and
\IEEEauthorblockN{4\textsuperscript{th} Morgan Chase McClellan}
\IEEEauthorblockA{\textit{Neuroscience Center} \\
\textit{Brigham Young University}\\
Provo, UT, USA \\
mchase11@byu.edu}
\and
\IEEEauthorblockN{5\textsuperscript{th} Dallan Gardner}
\IEEEauthorblockA{\textit{Department of Statistics} \\
\textit{Brigham Young University}\\
Provo, UT, USA \\
dg1254@byu.edu}
\and
\IEEEauthorblockN{6\textsuperscript{th} Dawson Hedges}
\IEEEauthorblockA{\textit{Neuroscience Center and Department of Psychology} \\
\textit{Brigham Young University}\\
Provo, UT, USA \\
dawson\_hedges@byu.edu}
}

\maketitle

\begin{abstract}
Manifold alignment (MA) involves a set of techniques for learning shared representations across domains, yet many traditional MA methods are incapable of performing out-of-sample extension, limiting their real-world applicability. We propose a guided representation learning framework leveraging a geometry-regularized twin autoencoder (AE) architecture to enhance MA while enabling generalization to unseen data. Our method enforces structured cross-modal mappings to maintain geometric fidelity in learned embeddings. By incorporating a pre-trained alignment model and a multitask learning formulation, we improve cross-domain generalization and representation robustness while maintaining alignment fidelity. We evaluate our approach using several MA methods, showing improvements in embedding consistency, information preservation, and cross-domain transfer. Additionally, we apply our framework to Alzheimer’s disease diagnosis, demonstrating its ability to integrate multi-modal patient data and enhance predictive accuracy in cases limited to a single domain by leveraging insights from the multi-modal problem.
\end{abstract}


%%%%%%%%%%%%%%%%%%%%%%%%%%%%%%%%%%%%%%%%%%%%%%%%%%%%%
\section{Introduction}\label{sec:intro}
%%%%%%%%%%%%%%%%%%%%%%%%%%%%%%%%%%%%%%%%%%%%%%%%%%%%%

Manifold learning encompasses a set of methods used to create a lower-dimensional representation, or an embedding, of higher-dimensional data. These embeddings highlight the relationships between intrinsic properties in the original data.  Such representations can form a key role in data visualization~\cite{tenenbaum2000isomap, van2008tsne, becht2019umap, moon2019phate, rhodes2023sampta}, dimensionality reduction as a preprocessing step for subsequent machine-learning or analytical tasks~\cite{duque2023dta}, or serve as a denoising mechanism~\cite{moon2019phate}. In the context of multi-domain problems, where multiple types of data are considered, manifold learning becomes more challenging as data distributions across different domains or modalities may exhibit domain-specific variations while still sharing a common geometric structure. 

Manifold alignment (MA) seeks to address this problem. In some contexts, a common, shared representation of multi-modal data can be viewed as a natural extension of manifold learning. For example, cell samples of the same type but collected at a different time or using different methodologies should still share features in common, but differences in the measured features may occur due to batch effects~\cite{cao2021ma-batch}, obscuring the similarities. In other contexts, such as language translations of the same document, we would expect extracted representations to share similar features when manifold learning is applied~\cite{wang2011manifold}.

A limitation of traditional manifold learning methods (e.g., $t$-SNE~\cite{van2008tsne}, Isomap~\cite{tenenbaum2000isomap}, Laplacian Eigenmaps~\cite{belkin2003laplacian}, Diffusion Maps~\cite{coifman2006dm})—and, by extension, related manifold alignment (MA) approaches~\cite{wang2011manifold, wang2011heterogeneous-da, lafon2006datafusion}—is the lack of a natural mechanism for out-of-sample extension, which hinders their ability to generalize to unseen data. When new data are introduced, the method must be re-run, in part or in full, and the new points—such as a test set—will influence the shape of the learned representation, limiting the use of this representation for subsequent machine learning tasks without the introduction of data leakage. This severely limits their utility in real-world applications, in particular, when at least one of the sourced domains is costly or difficult to attain.

Traditional methods for out-of-sample extension in manifold learning include kernel-based techniques, such as the Nyström extension~\cite{bengio2003out}, and landmark-based approximations~\cite{desilva2004sparse-mds, silva2005landmark-lasso, moon2019phate}, which estimate embeddings based on proximity to reference points. While kernel methods have been successfully integrated into eigenvector-based approaches like Isomap and Locally Linear Embedding (LLE)~\cite{bengio2003out}, they struggle with scalability and capturing complex nonlinear structures~\cite{aizenbud2015pca}, limiting their applicability to diffusion-based techniques and multi-modal alignment.

Neural networks offer a parametric alternative by learning embedding functions, as seen in parametric variants of $t$-SNE~\cite{van2009parametric-tsne} and UMAP~\cite{sainburg2021parametricumap}. However, direct regression onto the embedding space often leads to overfitting and poor generalization~\cite{arpit2017closer}, as the learned mappings may fail to preserve intrinsic geometric structures. To address this, multitask learning, such as that done by Autoencoders (AEs) have emerged as an effective regularization strategy to preserve geometric properties~\cite{zhang2018multi-task-learning, duque2020grae}. AE-based approaches, particularly geometry-regularized AEs~\cite{duque2020grae}, enhance generalization by jointly reconstructing input data while predicting embeddings, preserving essential geometric properties.

Building on this foundation, we propose a guided representation learning framework for MA with out-of-sample extension. Our approach employs a twin set of regularized AEs to preserve geometric fidelity in the learned embeddings while enforcing structured cross-modal mappings. By leveraging a pre-trained alignment model within a multitask learning formulation, we improve the robustness of representation learning, enhance cross-domain generalization, and increase alignment fidelity, refining and extending the original alignment process to better support generalization to unseen data.

\section{Related Works}\label{sec:related_works}

MA methods aim to learn shared representations across domains, often under semi-supervised settings with partial correspondences~\cite{ham2005ssma}. Classical approaches such as Joint Laplacian Manifold Alignment (JLMA)~\cite{wang2011manifold, singh2020unsup-align-omics}, which constructs a joint graph Laplacian across domains, and Manifold Alignment via Procrustes Analysis (MAPA)~\cite{wang2008ma-procrustes}, which applies Laplacian Eigenmaps followed by Procrustes alignment, rely on spectral embeddings. Later methods introduced variants using Diffusion Maps (DM)~\cite{lafon2006datafusion}, adversarial learning via the Manifold Alignment Generative Adversarial Network (MAGAN)~\cite{amodio2018magan}, or Diffusion Transport Alignment (DTA), which integrates diffusion processes with regularized optimal transport~\cite{duque2023dta}. More recent graph-based approaches such as Manifold Alignment via Stochastic Hopping (MASH) and Shortest Paths on the Union of Domains (SPUD)~\cite{rhodes2024gdi} preserve local and global structures through diffusion or shortest-path computations. These methods generally learn embeddings but lack principled mechanisms for out-of-sample extension.

Some MA methods (e.g., MAGAN, DTA, MASH) also enable cross-domain mapping, either through generators, barycentric projection, or diffusion operators. However, these mappings are tied to training data and do not naturally generalize to unseen samples.

Traditional manifold learning techniques, such as Isomap and LLE, can be extended to new data points using Nyström or landmark-based approximations~\cite{bengio2003out, desilva2004sparse-mds, silva2005landmark-lasso}. While computationally efficient, these methods scale poorly and often fail to capture complex nonlinear structures. Neural network–based extensions, including parametric $t$-SNE~\cite{van2009parametric-tsne} and UMAP~\cite{sainburg2021parametricumap}, offer a more flexible alternative but risk overfitting without additional structural constraints.

AE-based approaches, especially geometry-regularized variants~\cite{duque2020grae}, address these challenges by coupling reconstruction with embedding prediction, preserving geometric fidelity while improving generalization. Building on this foundation, we propose a guided twin AE framework that integrates pre-aligned embeddings in a multitask setting, enabling both alignment fidelity and robust out-of-sample extension. Unlike MAGAN, DTA, or MASH, which provide either limited or data-tied cross-domain mappings, our framework jointly enforces alignment fidelity and reconstruction through geometry-regularized twin AEs, thereby enabling scalable out-of-sample extension. More detailed descriptions of each of these approaches can be found in Appendix~\ref{sec:ext_related_works}.

%%%%%%%%%%%%%%%%%%%%%%%%%%%%%%%%%%%%%%%%%%%%%%%%%
\section{Methods}\label{sec:methods}
%%%%%%%%%%%%%%%%%%%%%%%%%%%%%%%%%%%%%%%%%%%%%%%%%

We use a geometry-regularized twin AE architecture for semi-supervised MA, extending the AE framework to a cross-modal mapping with dual regularization. Unlike traditional AEs that enforce self-preserving mappings, our approach employs a pair of networks regularized to maintain self-domain structure while enabling cross-modal translation. A key modification in our framework is the introduction of two regularization terms: one that aligns the bottleneck representations with a precomputed, aligned embedding, and another that guides the decoder network, taking advantage of known correspondences. Any semi-supervised alignment model can serve to generate the precomputed embeddings.

Here we introduce the notation used to describe the model. We note that our current work and subsequent experiments focus on two domains, but the idea is generalizable to $N > 2$ domains by introducing $N$ AEs. We defer experimentation on multiple domains to a later work.

Given two datasets, $X \subset \mathcal{X}$ and $Y \subset \mathcal{Y}$, we generate points to a shared latent space, $E \subset \mathcal{E}$ (e.g., $\mathcal{E} = \mathbb{R}^2$), using an embedding model of choice (see Appendix~\ref{subsec:manifold_alignment}). The encoded representations of $X$ and $Y$ are denoted respectively as $E_X and, E_Y \subset E$. The precomputed embeddings are used to guide the representation learning of the two encoders, ensuring that the mapping functions learn the pre-aligned embedding structure.

The architecture consists of two independent AEs, $AE_X$ and $AE_Y$, each comprising an encoder-decoder pair $(f_X, g_X)$ and $(f_Y, g_Y)$, respectively. The encoders $f_X$ and $f_Y$ map inputs $X$ and $Y$ to their latent representations, $E_X$ and $E_Y$, both belonging to the same, aligned space. That is, $f_X : X \to E$ such that $f_X(x_i) = e_{x_i} \in E_X$. The decoders $g_X$ and $g_Y$ reconstruct the inputs from these representations and are also used to map from the aligned embedded space to the respective feature spaces. 
% \begin{figure}[h]
%     \centering
%     \includegraphics[width=0.75\linewidth]{figures/Twin Autoencoders.png}
%     \caption{The twin autoencoder architecture. Both GRAE models are trained on the original data and its pre-aligned embedding.}
%     \label{fig:enter-label}
% \end{figure}
Each AE is trained independently using its respective dataset and embedded points from an existing alignment model. The independent training phase allows each AE to extract domain-specific features while preserving the shared latent structure. The decoder plays a role in guiding representation learning, ensuring that embeddings generalize effectively to unseen data within their respective domains while maintaining consistency in the shared latent space. After training, the decoders between the models can be swapped, enabling points from either domain to be transformed to the other. This is accomplished by embedding the points using the first AEs encoder and then reverse-transforming them back into the other domain using the other AEs decoder.

To guide the representation learning and ensure accurate mapping, we train the encoder functions with a regularized loss function such that (1) the embedded points are well-aligned in the low-dimensional space, preserving similarities between points across domains, and (2) the original data can be accurately reconstructed from its representation in the embedded space. Stricter alignment is enforced for known correspondences (anchors), ensuring that the points are well-aligned in the embedded space.

To these ends, the total loss function is defined as:

\[
\mathcal{L} = \mathcal{L}_{\text{recon}} + \lambda \mathcal{L}_{\text{align}} + \mathcal{L}_{\text{anchor}}
\]

where $\lambda$ is a hyperparameter controlling the influence of the geometric alignment. Initial experiments with the anchor loss showed that the influence of a coefficient (e.g., a $\lambda_{2}$) meant to control the influence of the anchor term had a negligible impact, as it quickly goes to 0 due to good alignment with the underlying MA model. To ensure the AEs effectively reconstruct the input data, we define the reconstruction loss as:

\[
\mathcal{L}_{\text{recon}} = \frac{1}{n_x} \sum_{i=1}^{n_x} \| \mathbf{x}_i - g_X(f_X(\mathbf{x}_i)) \|^2
\]

where $\mathbf{x}_i$ represents an input sample, and $f_X$ is the encoder function for domain $X$, $g_X$ is the corresponding decoder, and $n_x$ is the number of points in $X$.. To enforce alignment with the shared embedding space, we introduce an alignment loss term:

\[
\mathcal{L}_{\text{align}} = \frac{1}{n_x} \sum_{k=1}^{n_{x}} \| f_X(\mathbf{x}_k) - \mathbf{e}_{x_k} \|^2
\]

where $\mathbf{e}_{x_k}$ is the embedded point in $E$ corresponding to $x_k$. This term ensures that the encoded representations of the input datasets remain close to their aligned embeddings, thus improving the cross-domain consistency. Finally, we introduce the anchor loss term. Let $a_{x_i} \in Y$ denote anchor (known correspondence) of $x_i \in X$ (likewise, $a_{y_j} \in X$ being the known correspondence of $y_j \in Y$). Then

\[
\mathcal{L}_{\text{anchor}} = \frac{1}{n_A} \sum_{j=1}^{n_{A}} \| f_X(\mathbf{x}_j) - \mathbf{e}_{a_{x_j}} \|^2,
\]

where $n_A$ is the number of known correspondences. The anchor loss term enforces consistency between the embeddings' corresponding anchor points across domains, ensuring they map to the same latent representation. Specifically, for each designated anchor point $\mathbf{x}_j$ in one domain, its encoded representation $f_X(\mathbf{x}_j)$ is encouraged to align with the corresponding embedding $\mathbf{e}_{a_{x_j}}$ from the other domain. Since anchor points are shared across both domains, this constraint ensures that their representations remain consistent, reinforcing the cross-domain alignment and improving the overall structural coherence of the learned embedding space. This loss does not appear to change much during the training process, but keeping it as a part of the overall loss function prevents well-aligned known correspondences from drifting during training. By prioritizing structural constraints at the latent representation level, we enhance the model’s ability to learn a multi-domain embedding function that preserves the integrity of both modalities.

The work of~\cite{rhodes2024gdi} showed that typically 10\% of data as known correspondences is often sufficient for good alignment for most alignment methods, though some methods (DTA, JMLA) can perform well at lower percentages. As this work focuses on the extension and cross-domain mapping, rather than the initial alignment, we do not further explore the dependency of the anchors on these areas.

%%%%%%%%%%%%%%%%%%%%%%%%%%%%%%%%%%%%%%%%%%%%%%%%%
\subsection{Alignment Loss}
%%%%%%%%%%%%%%%%%%%%%%%%%%%%%%%%%%%%%%%%%%%%%%%%%

The alignment loss drives the network to map points with known correspondence to their corresponding points in the other domain. By ensuring that these anchor points are mapped to their correspondence, other nearby points without known correspondence are also properly positioned. $\lambda$ determines the strength of the requirement to map known correspondences. The larger the $\lambda$ value, the greater emphasis is placed on proper point mapping. 

We measure the precision of the relative positioning of the AE-mapped test points using Mantel's test~\cite{mantel1967detection}. Mantel’s test is a statistical test designed to assess the correlation between two distance (or dissimilarity) matrices. The test determines whether the structure of one distance matrix is significantly associated with the structure of another. In our context, we calculate the pairwise distances between test points (across both domains) in the embedding space, using the original alignment method ($D_{MA}$) and the AE model ($D_{AE}$), using the MA methods described in Appendix~\ref{subsec:manifold_alignment}. 

\begin{figure}[!htb]
    \centering
    \includegraphics[width=0.99\linewidth]{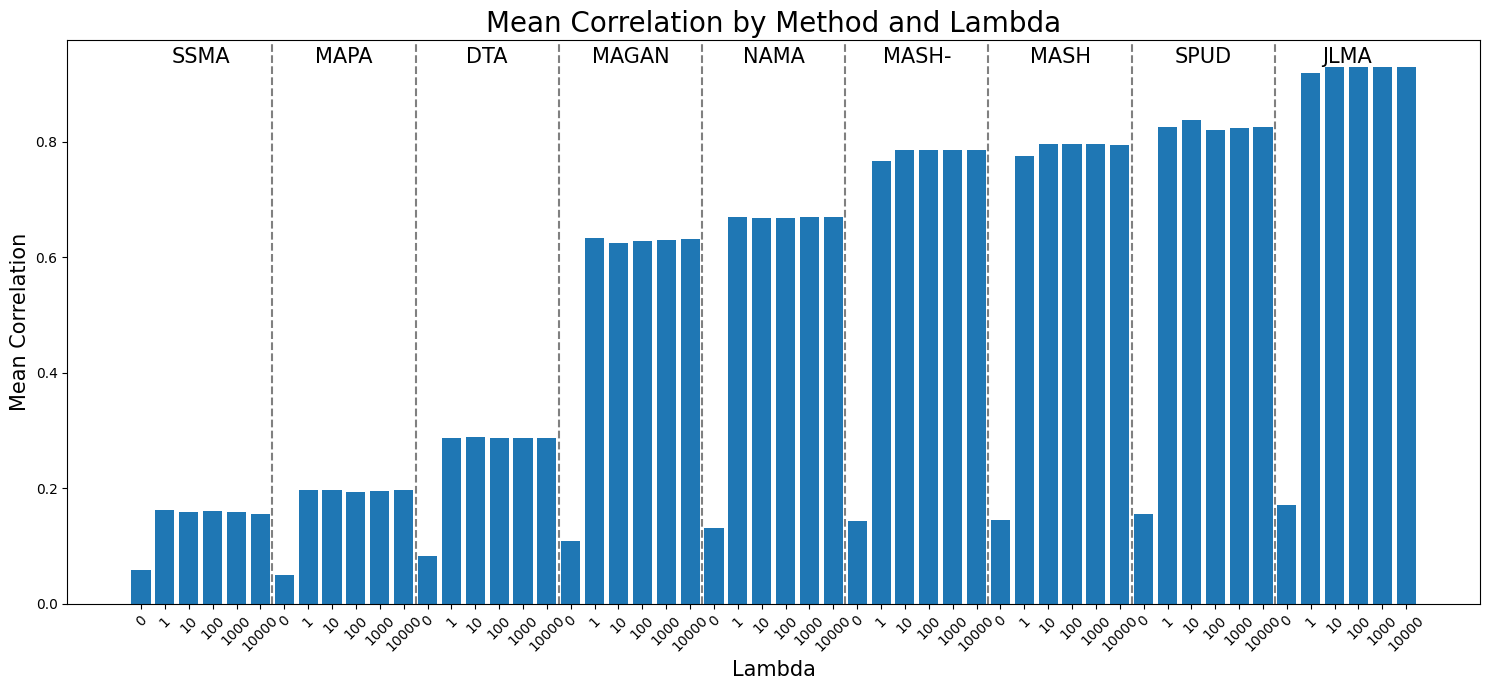}
    \caption{The Mantel correlations between the pairwise distances of the twin AE model aligned embedding and of those using the full, original embeddings using several $\lambda$ values. When $\lambda > 0$, the correlations are rather consistent within a given MA method. Without the alignment loss ($\lambda = 0$), the correlations are much smaller. All correlations are significant ($p < 0.01$), but they are much smaller for DTA, MAPA, and SSMA.}
    \label{fig:lambdas}
\end{figure}

Empirically, we see that the Mantel test scores are fairly consistent for all selected $\lambda > 0$. In Figure~\ref{fig:lambdas}, we see that within a given alignment method, $\lambda$ values of 1, 10, 100, 1,000, 10,000 all correspond to similar correlations, all of which are much higher than when $\lambda$ = 0. However, we also see that the Mantel correlations are not consistent among the MA methods used for regularization. The twin AE correlations are much lower for DTA, MAPA, and SSMA, and greatest for JLMA.

% TODO: Discuss more about the training/retraining process.

% \subsection{Correspondence Fine-Tuning}

% After this initial training, we can enhance cross-modal mapping by leveraging known correspondences to refine the decoding process. Specifically, we fine-tune the decoders to generate representations in domain $\mathcal{X}$ from embedded points in the set $E_Y$, and vice versa. %This process strengthens the alignment between the latent space and the corresponding data distributions.

% To achieve this, we refine each AE by (1) swapping the decoders $g_X$ and $g_Y$ within the AE structures and (2) encouraging the decoders to learn accurate cross-domain correspondences. To ensure accurate point reconstruction, the fine-tuning process of the decoders only uses known correspondences, minimizing the loss between the generated and actual correspondences of $x_k$. Specifically, we optimize:

% \[
% L(g_Y(e_{x_i}), a_{x_i})
% \]

% This approach reinforces the decoder’s ability to map between domains while preserving the underlying shared structure of the latent space.

%%%%%%%%%%%%%%%%%%%%%%%%%%%%%%%%%%%%%%%%%%%%%%%%%
\subsection{Extending to Single-Domain Points}
%%%%%%%%%%%%%%%%%%%%%%%%%%%%%%%%%%%%%%%%%%%%%%%%%

Acquiring data from different domains can vary significantly in cost and labor intensity. For example, gathering data through a user-filled survey is considerably less expensive than administering a full battery of tests designed to diagnose a specific disease. In many cases, the more costly domain tends to yield higher-quality data that is directly relevant to the practitioner’s task, such as supporting accurate diagnoses.

Within the framework of MA, the goal is to learn similarity measures between observations across different domains. Depending on the strength of the inter-domain relationships, it is possible to map data points from the inexpensive domain to those in the expensive domain by employing a pre-trained alignment model that leverages known connections between the two.

Using the twin AE architecture, new data points from the cost-effective domain can be processed to extract information corresponding to points in the expensive domain. This approach can improve performance in downstream tasks, such as prediction, by operating within the shared latent space.

Since the AEs are regularized using the shared embedding space, the test points extended to the embedding space can be subsequently mapped to their nearest representation in the other domain. Thus, we produce a mapping between feature spaces that has been guided using the known correspondences. More on this can be found in Appendix~\ref{sec:add_results}. Generally, the precision of the mapping depends on the number of known correspondences in the original training set and the ability of the underlying MA model to learn similarities between points of the different domains~\cite{rhodes2024gdi}. The code is available for the methods and experiments. See \href{https://github.com/JakeSRhodesLab/TwinAE-ManifoldAlignment}{https://github.com/JakeSRhodesLab/TwinAE-ManifoldAlignment}.
%%%%%%%%%%%%%%%%%%%%%%%%%%%%%%%%%%%%%%%%%%%%%%%%%%%%%
\section{Model Evaluation}
%%%%%%%%%%%%%%%%%%%%%%%%%%%%%%%%%%%%%%%%%%%%%%%%%%%%%

The evaluation of our twin AE models requires us to look at (1) the quality of the embedding mapping, that is, given a pretrained alignment model, how well the AE representations map to the aligned embedding space; (2) the quality of the information retained by the alignment model, as the AE is important to provide the out-of-sample extension required to map previously unseen data; (3) and the quality of the cross-domain mapping. Most MA methods do not provide a direct means to perform this task, but we compare the twin AE model against those that can. We discuss each of these evaluations in the following sections.

Unless otherwise stated, all datasets used in the below evaluations can be found in the UC Irvine Machine Learning Repository~\cite{dua2019uci}, \href{https://archive.ics.uci.edu/}{https://archive.ics.uci.edu/}. The specific datasets used are described in Tables~\ref{tab:class_datasets} and \ref{tab:reg_data} in the supplementary materials.

\subsection{Unimodal Domain Splits (Data Simulation)}

To simulate multi-domain data coming from a unimodal source, we employ feature-level domain splits or add different forms of distortion (e.g., random rotations) to simulate a second dataset. We briefly describe these here. More details about these simulations can be found in~\cite{rhodes2024gdi}.

\begin{enumerate}
    \item \textit{Random (random)} – Each domain is assigned a unique set of random features with no overlap.

    \item \textit{Skewed Importance (skewed)} – Features are ranked by importance using random forests. The top half forms domain $\mathcal{X}$, while the less relevant half forms $\mathcal{Y}$.

    \item \textit{Even Importance (even)} – Predictive features are split evenly between $\mathcal{X}$ and $\mathcal{Y}$, with less relevant features randomly assigned.

    \item \textit{Gaussian Noise (distort)} – Gaussian noise is added to all features, creating domain $\mathcal{Y}$.

    \item \textit{Random Rotation (rotation)} – Domain $\mathcal{Y}$ is generated by applying a random rotation to $\mathcal{X}$ using QR factorization.
\end{enumerate}

%%%%%%%%%%%%%%%%%%%%%%%%%%%%%%%%%%%%%%%%%%%%%%%%%%%%%
\subsection{AE Embedding Fit}
%%%%%%%%%%%%%%%%%%%%%%%%%%%%%%%%%%%%%%%%%%%%%%%%%%%%%

To evaluate the twin AE fit, we first work with the assumption that the alignment model provides the ground-truth values. The extent to which this assumption is met has been evaluated in other works~\cite{wang2011manifold, wang2008ma-procrustes, duque2023mali, duque2023dta, rhodes2024gdi}. For each dataset, we run each MA method on the full set (including training and test points). Then, using only the training data from each domain, we train our twin AEs. The AEs are then used to embed the test points. Given the ground truth of the full MA, we know where the test points should fall in the latent space, regardless of domain. Ideally, the relative locations of the embedded test points should be preserved using the AEs. We measure the distances between the relative locations using Mantel's test. See the results in Table~\ref{tab:mantels}. Here we see some dependence on the underlying MA model on the mapping abilities of the twin AE models. Some methods, such as JLMA, SPUD, MASH, have well-preserved embeddings, while MAGAN, DTA, and MAPA have embeddings that do not match the underlying aligned embeddings as closely.

\begin{table}[h]
    \centering
    \caption{\normalfont Mantel's test between model embedding distances and the twin AE-generated embedding distances, averaged across all test datasets and splits. All models exhibit significant correlation, indicating strong alignment between manifold-aligned embeddings and twin AE embeddings. However, the correlation is notably weaker for DTA ($r = 0.25$) and MAPA ($r = 0.17$), compared to others like JLMA ($r = 0.80$), SPUD ($r = 0.72$), and MASH ($r = 0.68$).}
    \scriptsize
    \renewcommand{\arraystretch}{1.2}
    \begin{tabular}{|l|c|c|}
        \hline
        \textbf{MA Method} & \textbf{Correlation} ${\pm}$ \textbf{SD} & \textbf{Significance Level} \\
        \hline
        JLMA  & $0.80 \pm 0.32$ & $0.006$ \\
        SPUD  & $0.72 \pm 0.29$ & $0.005$ \\
        MASH  & $0.68 \pm 0.33$ & $0.004$ \\
        % MASH- & $0.68 \pm 0.33$ & $0.005$ \\
        % NAMA  & $0.58 \pm 0.27$ & $0.009$ \\
        MAGAN & $0.54 \pm 0.23$ & $0.017$ \\
        DTA   & $0.25 \pm 0.12$ & $0.006$ \\
        MAPA  & $0.17 \pm 0.12$ & $0.014$ \\
        \hline
    \end{tabular}
    \vspace{1em}

    \label{tab:mantels}
\end{table}

%%%%%%%%%%%%%%%%%%%%%%%%%%%%%%%%%%%%%%%%%%%%%%%%%%%%%
\subsection{Model Baseline Comparisons (Information Quality)}
%%%%%%%%%%%%%%%%%%%%%%%%%%%%%%%%%%%%%%%%%%%%%%%%%%%%%

The embedding representations can be used for several purposes, such as multi-domain relationship visualizations~\cite{duque2023dta, rhodes2024gdi} (e.g., scatterplots of the aligned 2D embeddings), downstream prediction~\cite{rhodes2024RFMA, duque2023mali}, or exploratory processes, such as anomaly detection or clustering~\cite{singh2023unsup-man-align-single-cell, shaw2024forestproximitiestimeseries}. As an objective measure of alignment fit, we compare machine learning models trained on the MA embedding with models trained on each original domain. We use our AE networks to extend test points to the embedding space to prevent the test points from influencing the resulting embedding. 

Here is a simple description of this process: Given datasets $X \subset \mathcal{X}$ and $Y \subset \mathcal{Y}$, we train a machine learning model (random forest, $k$- Nearest Neighbors ($k$-NN) model) on each domain separately and evaluate their performance on respective test sets. Then, taking only the training data from each set, we run a MA method, resulting in an aligned embedding. The aligned embedding is used to regularize the twin AE models. After training the AE models, they are used to extend the test sets (from both domains) into the shared latent space. We then train the predictive models on the embedded, aligned training points from each domain and evaluate the models on their respective, mapped test points. Ideally, the aligned representation should include enough information to enhance the predictive model performance in both domains. However, we are content to see the performance enhanced for at least one of the domains, thus suggesting that the domain with richer information assists the domain with poorer information. This is helpful in real-world applications where domains with richer information are more expensive or difficult to attain.

\begin{figure}
    \centering
    \includegraphics[width=\linewidth]{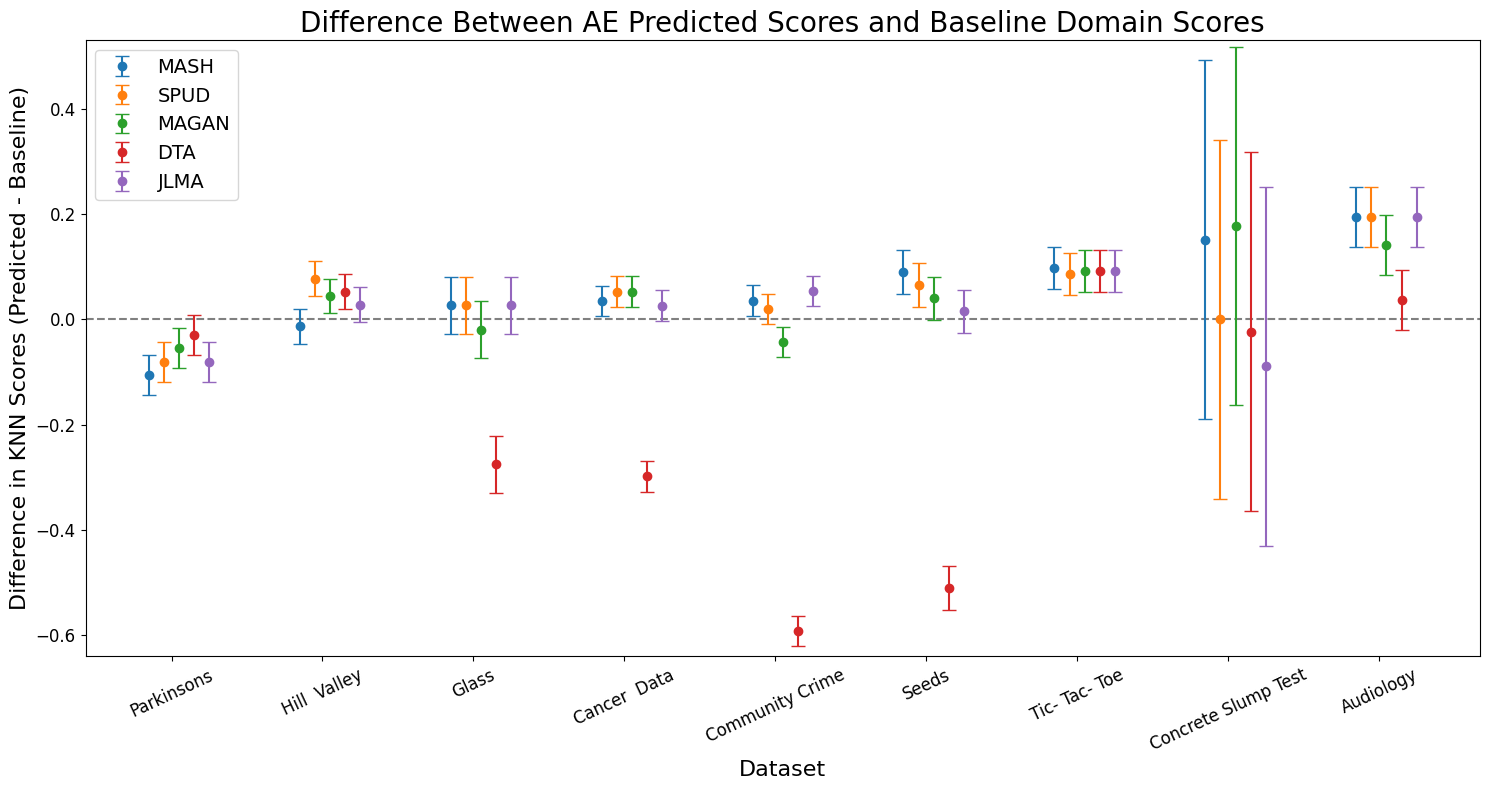}
    \caption{$k$-Nearest Neighbors ($k$-NN) was first evaluated on the weaker domain to establish baseline performance. Both the stronger and weaker domains were then processed using the given manifold alignment methods. A new $k$-NN model was trained on the resulting aligned embedding. Additionally, test points from the weaker domain were mapped into the shared domain space via the twin AE structure for prediction. Each experiment was conducted across five different random seeds. The datasets were chosen to allow separation into distinct weaker and stronger feature subsets. Results indicate that MASH, SPUD, JLMA, and MAGAN consistently improve performance, whereas DTA often exhibits instability.}
    \label{fig:baseline-comparisons}
\end{figure}

In Figure~\ref{fig:baseline-comparisons}, we compare the baseline performance of a $k$-NN model on the weaker domains to the model trained on the aligned, training embedding from the AE models, using MASH, SPUD, MAGAN, and DTA for regularization. These include all MA models with the highest Mantel correlations and DTA for comparison with a method with lower Mantel correlations. For all but one dataset, the AE-aligned model outperformed the baseline with at least two of the MA methods. Of the MA regularizations, DTA had the least consistent results, while improvement was more consistent with the other regularizations.

%%%%%%%%%%%%%%%%%%%%%%%%%%%%%%%%%%%%%%%%%%%%%%%%%%%%%
\subsection{Cross-Domain Mapping}\label{sec:cross-domain-map}
%%%%%%%%%%%%%%%%%%%%%%%%%%%%%%%%%%%%%%%%%%%%%%%%%%%%%

A minority of MA methods are capable of providing a direct map between domains. Most provide a mechanism to learn some form of cross-domain similarity measure, which is subsequently used with an embedding function (e.g., Laplacian Eigenmaps~\cite{wang2008ma-procrustes, singh2020unsup-align-omics}, Diffusion Maps~\cite{lafon2006datafusion}, or multidimensional scaling (MDS)~\cite{shen2017manifold-matching, rhodes2024gdi}). DTA, MASH, and MAGAN methods provide an inter-domain mapping function that we compare with our AE cross-domain reconstructions.

DTA~\cite{duque2023dta} learns a coupling matrix via entropic optimal transport that can serve as a projection matrix for cross-domain adaptation. The learned coupling matrix serves as a function to map each point in $X$ to its nearest representation in $Y$, essentially shuffling the points according to the coupling matrix. This is not a general function capable of mapping new points to the opposing space but can be used as a mechanism to assess the consistency of the alignment. 

MASH~\cite{rhodes2024gdi} learns a cross-domain diffusion operator that, when row-normalized, serves as a set of weights that can be used to project representations of points in $X$ as points in $Y$ as a weighted sum. The projected points from $X$ using MASH are not guaranteed to map to actual points in $Y$ but are newly generated points that come from a similar distribution, that is, mapped points share common distributional central tendencies. A practical use for this methodology is to make inferences about other source domains using information learned from multiple domains, as is demonstrated in Section~\ref{sec:alz}.

MAGAN~\cite{amodio2018magan} is a Generative Adversarial Network that is naturally intended for data generation. The generator models are a direct mapping function from $X$ to $Y$ and vice versa.

We evaluate cross-domain mapping accuracy by comparing predicted points to their true correspondences. Across datasets, the decoder-swapping mechanism of our twin AEs consistently yields lower errors than competing methods (MAGAN, DTA, and MASH), regardless of the underlying alignment model. Full results can be found in Appendix~\ref{sec:add_results}.

%%%%%%%%%%%%%%%%%%%%%%%%%%%%%%%%%%%%%%%%%%%%%%%%%%%%%
\section{Alzheimer's Case}\label{sec:alz}
%%%%%%%%%%%%%%%%%%%%%%%%%%%%%%%%%%%%%%%%%%%%%%%%%%%%%

Alzheimer’s disease (AD) is the most common form of dementia, accounting for 60--70\% of cases~\cite{Alzheimer'sAssociation2024, WHO2023}. In the U.S., prevalence among adults over 65 is estimated at 14\% for African Americans, 12\% for Hispanics, and 10\% for non-Hispanic whites~\cite{Lim2023}. Early diagnosis is challenging due to a preclinical, asymptomatic phase, where symptoms may resemble normal aging. While no cure currently exists, symptom management can help maintain quality of life, particularly with early intervention and attention to individual needs~\cite{Alzheimer'sAssociation2024, WHO2023}.

Common assessment tools include the ADAS-Cog 13~\cite{Rosen1984}, which evaluates cognitive functions like memory and language through a battery of 13 clinical tests~\cite{Clarke2022}, and the Functional Activities Questionnaire (FAQ), which assesses ability to perform 10 essential daily tasks~\cite{Gonzalez2022}. Both provide insight into different aspects of AD progression (see Tables~\ref{tab:ADAS_Variables} and~\ref{tab:FAQ_Variables}). 

Using data from the Alzheimer’s Disease Neuroimaging Initiative (ADNI), we combine and translate between these tests to better understand the relationship between cognitive and functional decline, especially when only one measure is available. ADNI is a longitudinal study launched in 2004, involving over 3{,}000 patients across five cohorts. For this work, we used visit histories from the ADNI3 cohort (approximately 1{,}200 patients). ADAS-Cog 13 and FAQ served as data domains, and the clinician-assigned diagnosis at the time of the tests was the prediction target. We applied MASH manifold alignment using random forest proximities~\cite{rhodes2024RFMA}, trained twin AEs, and validated alignment performance as shown in Table~\ref{tab:accuracy_comparisons}.

\begin{table}
    \centering
    \scriptsize
    \caption{These values represent the raw accuracy of the ADAS-Cog 13 and FAQ scores as evaluated on the test set using a random forest model with 100 trees compared to the accuracy of the twin AE extension as evaluated by embedding the same test points and running a $k$-NN model to predict their labels. The extension accuracy is roughly analogous to the original accuracy.}

    \begin{tabular}{|l|c|c|} \hline 
        
         Domain&  Original Accuracy&  Extension Accuracy\\ \hline  
         ADAS-Cog 13&  63.0\%&  62.6\% \\ \hline 
         FAQ&  70.6\%&  66.1\% \\ \hline
         
    \end{tabular}

\label{tab:accuracy_comparisons}
\end{table}

The twin AE architecture was then utilized to translate data from the ADAS-Cog 13 domain to the FAQ domain and vice versa, via embedding the test data and then letting the opposite AE translate the embedded points into the other domain. The aggregated results of these translations are shown in Table~\ref{tab:translation_scores}. Individualized results can be found in Appendix~\ref{sec:adni_vars}.

\begin{table}
    \centering
    \scriptsize
    \caption{Each domain's data is translated to the other by either AE first embedding the points, and then reversing the embedding process using the twin AE to the other domain. The average RMSE between the translated scores to the actual scores on the opposite test are displayed. Possible scorings ranged from 5-12 points per question, see Supplemental Tables \ref{tab:FAQ_Variables} and \ref{tab:ADAS_Variables}.}
    \begin{tabular}{|c|c|c|} \hline 
         Origin&  Translated To& RMSE\\ \hline 
         FAQ&  ADAS-Cog 13& 0.899\\ \hline 
         ADAS-Cog 13&  FAQ& 1.040\\ \hline
    \end{tabular}
    \vspace{1em}

    \label{tab:translation_scores}
\end{table}
The ADAS-Cog to FAQ translation could be used practically to predict the difficulty that an AD patient might have in different areas of their life. Data points in the FAQ domain define the patient's ability to do things like ``Writing checks, paying bills, or balancing a checkbook", ``Shopping alone for clothes, household necessities, or groceries", and ``Remembering appointments, family occasions, holidays, medications" (See Supplemental Table~\ref{tab:FAQ_Variables}). The translation here could be useful in predicting what level and type of care an individual patient might require.

\section{Conclusion}\label{sec:conclusion}

% In this work, we introduced a geometry-regularized twin AE framework to address the challenges of manifold alignment with out-of-sample extension. Our approach integrates structured cross-modal mappings and pre-trained alignment models to enhance generalization, preserve geometric fidelity, and enable cross-domain transfer learning. Through empirical evaluations, we demonstrated that our method effectively maps out-of-sample points to meaningful points in the embedding space and is capable of providing meaningful cross-domain mappings. 

% We showcased a real-world application in Alzheimer’s disease (AD) cognitive testing, where our model successfully translated multi-modal patient assessments (ADAS-Cog 13 and FAQ scores), offering a practical tool for predicting functional impairments from cognitive evaluations. This case study underscores the potential of our method in domains where aligned multimodal data can enhance predictive accuracy.

In this work, we introduced a geometry-regularized twin AE framework to address the limitations of MA, extending to out-of-sample points. By integrating pre-aligned embeddings into a multitask learning structure, our model simultaneously preserves domain-specific information, enforces alignment fidelity, and helps generalization to unseen data. This framework enables both embedding construction and cross-domain translation, offering a flexible and scalable solution for multi-modal data alignment.

Empirical evaluations across a diverse set of benchmark datasets demonstrated that our approach reliably reconstructs aligned embeddings and supports meaningful cross-domain mappings, outperforming existing alternatives such as DTA, MAGAN, and MASH. Our experiments highlight the adaptability of the model to various underlying alignment methods and its ability to enhance predictive accuracy through cross-domain information transfer.

We also demonstrated our method through a real-world application in Alzheimer’s disease assessment, where it successfully translated between cognitive and functional evaluation scores. This demonstrates its practical potential in clinical settings where data availability across modalities may be imbalanced.

\section*{Acknowledgement}

The work presented in this article was in part funded by Simmons Research Endowment.

{
    \small
    \bibliographystyle{IEEEtran}
    \bibliography{main}

\begin{thebibliography}{10}
\providecommand{\url}[1]{#1}
\csname url@samestyle\endcsname
\providecommand{\newblock}{\relax}
\providecommand{\bibinfo}[2]{#2}
\providecommand{\BIBentrySTDinterwordspacing}{\spaceskip=0pt\relax}
\providecommand{\BIBentryALTinterwordstretchfactor}{4}
\providecommand{\BIBentryALTinterwordspacing}{\spaceskip=\fontdimen2\font plus
\BIBentryALTinterwordstretchfactor\fontdimen3\font minus \fontdimen4\font\relax}
\providecommand{\BIBforeignlanguage}[2]{{%
\expandafter\ifx\csname l@#1\endcsname\relax
\typeout{** WARNING: IEEEtran.bst: No hyphenation pattern has been}%
\typeout{** loaded for the language `#1'. Using the pattern for}%
\typeout{** the default language instead.}%
\else
\language=\csname l@#1\endcsname
\fi
#2}}
\providecommand{\BIBdecl}{\relax}
\BIBdecl

\bibitem{tenenbaum2000isomap}
\BIBentryALTinterwordspacing
J.~B. Tenenbaum, V.~Silva, and J.~C. Langford, ``A global geometric framework for nonlinear dimensionality reduction,'' \emph{Science}, vol. 290, no. 5500, pp. 2319--2323, 2000. [Online]. Available: \url{https://doi.org/10.1126/science.290.5500.2319}
\BIBentrySTDinterwordspacing

\bibitem{van2008tsne}
L.~Van~der Maaten and G.~Hinton, ``Visualizing data using t-sne.'' \emph{Journal of machine learning research}, vol.~9, no.~11, 2008.

\bibitem{becht2019umap}
E.~Becht, L.~McInnes, J.~Healy, C.-A. Dutertre, I.~W. Kwok, L.~G. Ng, F.~Ginhoux, and E.~W. Newell, ``Dimensionality reduction for visualizing single-cell data using umap,'' \emph{Nature biotechnology}, vol.~37, no.~1, pp. 38--44, 2019.

\bibitem{moon2019phate}
\BIBentryALTinterwordspacing
K.~R. Moon, D.~van Dijk \emph{et~al.}, ``Visualizing structure and transitions in high-dimensional biological data,'' \emph{Nat. Biotechnol.}, vol.~37, no.~12, pp. 1482--1492, Dec 2019. [Online]. Available: \url{https://doi.org/10.1038/s41587-019-0336-3}
\BIBentrySTDinterwordspacing

\bibitem{rhodes2023sampta}
J.~S. Rhodes, ``Supervised manifold learning via random forest geometry-preserving proximities*,'' in \emph{2023 International Conference on Sampling Theory and Applications (SampTA)}, 2023, pp. 1--5.

\bibitem{duque2023dta}
A.~F. Duque, G.~Wolf, and K.~R. Moon, ``Diffusion transport alignment,'' in \emph{Advances in Intelligent Data Analysis XXI}, B.~Cr{\'e}milleux, S.~Hess, and S.~Nijssen, Eds.\hskip 1em plus 0.5em minus 0.4em\relax Cham: Springer Nature Switzerland, 2023, pp. 116--129.

\bibitem{cao2021ma-batch}
\BIBentryALTinterwordspacing
K.~Cao, Y.~Hong, and L.~Wan, ``Manifold alignment for heterogeneous single-cell multi-omics data integration using pamona,'' \emph{Bioinformatics}, vol.~38, no.~1, pp. 211--219, 08 2021. [Online]. Available: \url{https://doi.org/10.1093/bioinformatics/btab594}
\BIBentrySTDinterwordspacing

\bibitem{wang2011manifold}
C.~Wang, P.~Krafft, and S.~Mahadevan, ``Manifold alignment,'' in \emph{Manifold Learning: Theory and Applications}, Y.~Ma and Y.~Fu, Eds.\hskip 1em plus 0.5em minus 0.4em\relax CRC Press, 2011.

\bibitem{belkin2003laplacian}
M.~Belkin and P.~Niyogi, ``Laplacian eigenmaps for dimensionality reduction and data representation,'' \emph{Neural computation}, vol.~15, no.~6, pp. 1373--1396, 2003.

\bibitem{coifman2006dm}
\BIBentryALTinterwordspacing
R.~R. Coifman and S.~Lafon, ``Diffusion maps,'' \emph{Appl. Comput. Harmon. Anal.}, vol.~21, no.~1, pp. 5--30, 2006, special Issue: Diffusion Maps and Wavelets. [Online]. Available: \url{https://doi.org/10.1016/j.acha.2006.04.006}
\BIBentrySTDinterwordspacing

\bibitem{wang2011heterogeneous-da}
C.~Wang and S.~Mahadevan, ``Heterogeneous domain adaptation using manifold alignment,'' in \emph{International Joint Conference on Artificial Intelligence}, 2011.

\bibitem{lafon2006datafusion}
S.~Lafon, Y.~Keller, and R.~Coifman, ``Data fusion and multicue data matching by diffusion maps,'' \emph{IEEE Transactions on Pattern Analysis and Machine Intelligence}, vol.~28, no.~11, pp. 1784--1797, 2006.

\bibitem{bengio2003out}
Y.~Bengio, J.~Paiement \emph{et~al.}, ``Out-of-sample extensions for lle, isomap, mds, eigenmaps, and spectral clustering,'' \emph{NeurIPS}, vol.~16, 2003.

\bibitem{desilva2004sparse-mds}
V.~De~Silva and J.~B. Tenenbaum, ``Sparse multidimensional scaling using landmark points,'' technical report, Stanford University, Tech. Rep., 2004.

\bibitem{silva2005landmark-lasso}
\BIBentryALTinterwordspacing
J.~Silva, J.~Marques, and J.~a. Lemos, ``Selecting landmark points for sparse manifold learning,'' in \emph{Advances in Neural Information Processing Systems}, Y.~Weiss, B.~Sch\"{o}lkopf, and J.~Platt, Eds., vol.~18.\hskip 1em plus 0.5em minus 0.4em\relax MIT Press, 2005. [Online]. Available: \url{https://proceedings.neurips.cc/paper_files/paper/2005/file/780965ae22ea6aee11935f3fb73da841-Paper.pdf}
\BIBentrySTDinterwordspacing

\bibitem{aizenbud2015pca}
Y.~Aizenbud, A.~Bermanis, and A.~Averbuch, ``Pca-based out-of-sample extension for dimensionality reduction,'' \emph{arXiv preprint arXiv:1511.00831}, 2015.

\bibitem{van2009parametric-tsne}
L.~Van Der~Maaten, ``Learning a parametric embedding by preserving local structure,'' in \emph{Artificial intelligence and statistics}.\hskip 1em plus 0.5em minus 0.4em\relax PMLR, 2009, pp. 384--391.

\bibitem{sainburg2021parametricumap}
T.~Sainburg, L.~McInnes, and T.~Q. Gentner, ``Parametric umap embeddings for representation and semisupervised learning,'' \emph{Neural Computation}, vol.~33, no.~11, pp. 2881--2907, 2021.

\bibitem{arpit2017closer}
D.~Arpit, S.~Jastrzebski, N.~Ballas, D.~Krueger, E.~Bengio, M.~S. Kanwal, T.~Maharaj, A.~Fischer, A.~Courville, Y.~Bengio \emph{et~al.}, ``A closer look at memorization in deep networks,'' in \emph{International conference on machine learning}.\hskip 1em plus 0.5em minus 0.4em\relax PMLR, 2017, pp. 233--242.

\bibitem{zhang2018multi-task-learning}
\BIBentryALTinterwordspacing
Y.~Zhang and Q.~Yang, ``An overview of multi-task learning,'' \emph{National Science Review}, vol.~5, no.~1, pp. 30--43, 09 2017. [Online]. Available: \url{https://doi.org/10.1093/nsr/nwx105}
\BIBentrySTDinterwordspacing

\bibitem{duque2020grae}
\BIBentryALTinterwordspacing
A.~F. Duque, S.~Morin, G.~Wolf, and K.~R. Moon, ``Extendable and invertible manifold learning with geometry regularized autoencoders,'' \emph{2020 IEEE International Conference on Big Data (Big Data)}, pp. 5027--5036, 2020. [Online]. Available: \url{http://doi.org/10.1109/BigData50022.2020.9378049}
\BIBentrySTDinterwordspacing

\bibitem{ham2005ssma}
\BIBentryALTinterwordspacing
J.~Ham, D.~Lee, and L.~Saul, ``Semisupervised alignment of manifolds,'' in \emph{Proceedings of the Tenth International Workshop on Artificial Intelligence and Statistics}, ser. Proceedings of Machine Learning Research, R.~G. Cowell and Z.~Ghahramani, Eds., vol.~R5.\hskip 1em plus 0.5em minus 0.4em\relax PMLR, 06--08 Jan 2005, pp. 120--127, reissued by PMLR on 30 March 2021. [Online]. Available: \url{https://proceedings.mlr.press/r5/ham05a.html}
\BIBentrySTDinterwordspacing

\bibitem{singh2020unsup-align-omics}
R.~Singh, P.~Demetci, G.~Bonora, V.~Ramani, C.~Lee, H.~Fang, Z.~Duan, X.~Deng, J.~Shendure, C.~Disteche, and W.~S. Noble, ``Unsupervised manifold alignment for single-cell multi-omics data,'' \emph{ACM BCB}, vol. 2020, pp. 1--10, Sep 2020, pubMed-not-MEDLINE.

\bibitem{wang2008ma-procrustes}
\BIBentryALTinterwordspacing
C.~Wang and S.~Mahadevan, ``Manifold alignment using procrustes analysis,'' in \emph{Proceedings of the 25th International Conference on Machine Learning}, ser. ICML '08.\hskip 1em plus 0.5em minus 0.4em\relax New York, NY, USA: Association for Computing Machinery, 2008, p. 1120–1127. [Online]. Available: \url{https://doi.org/10.1145/1390156.1390297}
\BIBentrySTDinterwordspacing

\bibitem{amodio2018magan}
\BIBentryALTinterwordspacing
M.~Amodio and S.~Krishnaswamy, ``Magan: Aligning biological manifolds,'' in \emph{International Conference on Machine Learning}, 2018. [Online]. Available: \url{https://api.semanticscholar.org/CorpusID:3303339}
\BIBentrySTDinterwordspacing

\bibitem{rhodes2024gdi}
J.~S. Rhodes and A.~G. Rustad, ``Graph integration for diffusion-based manifold alignment,'' in \emph{2024 International Conference on Machine Learning and Applications (ICMLA)}, 2024, pp. 1--8.

\bibitem{mantel1967detection}
N.~Mantel, ``The detection of disease clustering and a generalized regression approach,'' \emph{Cancer research}, vol.~27, no. 2\_Part\_1, pp. 209--220, 1967.

\bibitem{dua2019uci}
\BIBentryALTinterwordspacing
D.~Dua and C.~Graff, ``Uci machine learning repository,'' 2017. [Online]. Available: \url{http://archive.ics.uci.edu/ml}
\BIBentrySTDinterwordspacing

\bibitem{duque2023mali}
A.~F. Duque~Correa, M.~Lizotte, G.~Wolf, and K.~R. Moon, ``Manifold alignment with label information,'' in \emph{2023 International Conference on Sampling Theory and Applications (SampTA)}, 2023, pp. 1--6.

\bibitem{rhodes2024RFMA}
\BIBentryALTinterwordspacing
J.~S. Rhodes and A.~G. Rustad, ``{ Random Forest-Supervised Manifold Alignment },'' in \emph{2024 IEEE International Conference on Big Data (BigData)}.\hskip 1em plus 0.5em minus 0.4em\relax Los Alamitos, CA, USA: IEEE Computer Society, Dec. 2024, pp. 3309--3312. [Online]. Available: \url{https://doi.ieeecomputersociety.org/10.1109/BigData62323.2024.10825663}
\BIBentrySTDinterwordspacing

\bibitem{singh2023unsup-man-align-single-cell}
\BIBentryALTinterwordspacing
A.~Singh, K.~Biharie, M.~J.~T. Reinders, A.~Mahfouz, and T.~Abdelaal, ``sctopogan: unsupervised manifold alignment of single-cell data,'' \emph{Bioinformatics Advances}, vol.~3, no.~1, p. vbad171, 11 2023. [Online]. Available: \url{https://doi.org/10.1093/bioadv/vbad171}
\BIBentrySTDinterwordspacing

\bibitem{shaw2024forestproximitiestimeseries}
\BIBentryALTinterwordspacing
B.~Shaw, J.~Rhodes, S.~F. Boubrahimi, and K.~R. Moon, ``Forest proximities for time series,'' 2024. [Online]. Available: \url{https://arxiv.org/abs/2410.03098}
\BIBentrySTDinterwordspacing

\bibitem{shen2017manifold-matching}
\BIBentryALTinterwordspacing
C.~Shen, J.~T. Vogelstein, and C.~E. Priebe, ``Manifold matching using shortest-path distance and joint neighborhood selection,'' \emph{Pattern Recognition Letters}, vol.~92, pp. 41--48, 2017. [Online]. Available: \url{https://www.sciencedirect.com/science/article/pii/S016786551730106X}
\BIBentrySTDinterwordspacing

\bibitem{Alzheimer'sAssociation2024}
A.~Association, ``2024 alzheimer's disease facts and figures,'' \emph{Alzheimer's \& Dementia}, vol.~20, no.~5, 2024.

\bibitem{WHO2023}
\BIBentryALTinterwordspacing
W.~H. Organization. (2023) Dementia. [Online]. Available: \url{https://www.who.int/news-room/fact-sheets/detail/dementia#:~:text=Alzheimer%20disease%20is%20the%20most,dependency%20among%20older%20people%20globally}
\BIBentrySTDinterwordspacing

\bibitem{Lim2023}
A.~C. Lim, L.~L. Barnes, G.~H. Weissberger, M.~Lamar, A.~L. Nguyen, L.~Fenton, J.~Herrera, and S.~D. Han, ``Quantification of race/ethnicity representation in alzheimer’s disease neuroimaging research in the usa: a systematic review,'' \emph{Communications medicine}, vol.~3, no.~1, p. 101, 2023.

\bibitem{Rosen1984}
W.~G. Rosen, R.~C. Mohs, and K.~L. Davis, ``A new rating scale for alzheimer's disease.'' \emph{The American journal of psychiatry}, vol. 141, no.~11, pp. 1356--1364, 1984.

\bibitem{Clarke2022}
A.~Clarke, C.~Ashe, J.~Jenkinson, O.~Rowe, A.~, P.~Hyland, and S.~Commins, ``Predicting conversion of patients with mild cognitive impairment to alzheimer’s disease using bedside cognitive assessments,'' \emph{Journal of Clinical and Experimental Neuropsychology}, vol.~44, no.~10, pp. 703--712, 2022.

\bibitem{Gonzalez2022}
D.~A. Gonz{\'a}lez, M.~M. Gonzales, Z.~J. Resch, A.~C. Sullivan, and J.~R. Soble, ``Comprehensive evaluation of the functional activities questionnaire (faq) and its reliability and validity,'' \emph{Assessment}, vol.~29, no.~4, pp. 748--763, 2022.

\bibitem{gower2004procrustes}
J.~C. Gower and G.~B. Dijksterhuis, \emph{Procrustes problems}.\hskip 1em plus 0.5em minus 0.4em\relax OUP Oxford, 2004, vol.~30.

\bibitem{fitzgibbon2003icp}
\BIBentryALTinterwordspacing
A.~W. Fitzgibbon, ``Robust registration of 2d and 3d point sets,'' \emph{Image Vis. Comput.}, vol.~21, pp. 1145--1153, 2003. [Online]. Available: \url{https://api.semanticscholar.org/CorpusID:7576794}
\BIBentrySTDinterwordspacing

\bibitem{wofson1997geom-hashing}
H.~Wolfson and I.~Rigoutsos, ``Geometric hashing: an overview,'' \emph{IEEE Computational Science and Engineering}, vol.~4, no.~4, pp. 10--21, 1997.

\bibitem{villani2009optimaltransport}
C.~Villani \emph{et~al.}, \emph{Optimal transport: old and new}.\hskip 1em plus 0.5em minus 0.4em\relax Springer, 2009, vol. 338.

\bibitem{courty2014domain-reg-optimal}
N.~Courty, R.~Flamary, and D.~Tuia, ``Domain adaptation with regularized optimal transport,'' in \emph{Machine Learning and Knowledge Discovery in Databases: European Conference, ECML PKDD 2014, Nancy, France, September 15-19, 2014. Proceedings, Part I 14}.\hskip 1em plus 0.5em minus 0.4em\relax Springer, 2014, pp. 274--289.

\bibitem{peng2024scalablemanifoldlearninguniform}
\BIBentryALTinterwordspacing
D.~Peng, Z.~Gui, W.~Wei, and H.~Wu, ``Scalable manifold learning by uniform landmark sampling and constrained locally linear embedding,'' 2024. [Online]. Available: \url{https://arxiv.org/abs/2401.01100}
\BIBentrySTDinterwordspacing

\bibitem{Gigante2019CompressedD}
\BIBentryALTinterwordspacing
S.~A. Gigante, J.~S. Stanley, N.~Vu, D.~van Dijk, K.~R. Moon, G.~Wolf, and S.~Krishnaswamy, ``Compressed diffusion,'' \emph{2019 13th International conference on Sampling Theory and Applications (SampTA)}, pp. 1--4, 2019. [Online]. Available: \url{https://api.semanticscholar.org/CorpusID:59553282}
\BIBentrySTDinterwordspacing

\bibitem{roman-rangle2019inductive-tsne}
\BIBentryALTinterwordspacing
E.~Roman-Rangel and S.~Marchand-Maillet, ``Inductive t-sne via deep learning to visualize multi-label images,'' \emph{Engineering Applications of Artificial Intelligence}, vol.~81, pp. 336--345, 2019. [Online]. Available: \url{https://www.sciencedirect.com/science/article/pii/S0952197619300156}
\BIBentrySTDinterwordspacing

\bibitem{Le2018supervised_autoencoders}
\BIBentryALTinterwordspacing
L.~Le, A.~Patterson, and M.~White, ``Supervised autoencoders: Improving generalization performance with unsupervised regularizers,'' in \emph{Advances in Neural Information Processing Systems}, S.~Bengio, H.~Wallach, H.~Larochelle, K.~Grauman, N.~Cesa-Bianchi, and R.~Garnett, Eds., vol.~31.\hskip 1em plus 0.5em minus 0.4em\relax Curran Associates, Inc., 2018. [Online]. Available: \url{https://proceedings.neurips.cc/paper_files/paper/2018/file/2a38a4a9316c49e5a833517c45d31070-Paper.pdf}
\BIBentrySTDinterwordspacing

\end{thebibliography}
}

\newpage

\appendices

{\footnotesize

\section{Extended Related Works}\label{sec:ext_related_works}

In Appendix~\ref{subsec:manifold_alignment}, we discuss advances in manifold alignment generally. Besides neural-network-based approaches, none of the described methods have built-in means to extend to new points. A brief discussion on cross-domain mapping is described in Appendix~\ref{subsec:cross_domain_mapping}. In Appendix~\ref{subsec:manifold_learning_extensions}, we discuss approaches for out-of-sample extension in manifold learning, motivating our approach to extend embedded points in the alignment problem.  

%%%%%%%%%%%%%%%%%%%%%%%%%%%%%%%%%%%%%%%%%%%%%%%%%%%%%
\subsection{Manifold Alignment Methods}\label{subsec:manifold_alignment}
%%%%%%%%%%%%%%%%%%%%%%%%%%%%%%%%%%%%%%%%%%%%%%%%%%%%%

MA techniques extract data representations from two or more modalities or co-domains, although our work centers on cases of only two domains, denoted here as $\mathcal{X}$ and $\mathcal{Y}$. We limit our discussion of related works to semi-supervised approaches in which a partial but not a full correspondence is available between points in different domains, as described in~\cite{ham2005ssma}.

In the work of~\cite{ham2005ssma}, the authors introduced two semi-supervised MA methods. The first method relies on predefined embedding coordinates to constrain the mapping functions from both domains. The second approach, which aligns with the scenarios considered in this paper, uses partially annotated correspondences to preserve them in a low-dimensional space. Specifically, the cost function for Laplacian Eigenmaps is modified to minimize the distance between embedding coordinates of known correspondences. A similar method was introduced in~\cite{wang2011manifold}, where the authors proposed a method, later dubbed Joint Laplacian Manifold Alignment (JLMA)~\cite{singh2020unsup-align-omics}. JLMA employs a joint graph Laplacian constructed from domain-specific similarity matrices ($W_{\mathcal{X}}$ and $W_{\mathcal{Y}}$) and inter-domain similarities based on known correspondences. Laplacian Eigenmaps are applied to the joint Laplacian to produce a unified embedding.

The same authors introduced Manifold Alignment via Procrustes Analysis (MAPA)~\cite{wang2008ma-procrustes}, in which Laplacian Eigenmaps~\cite{belkin2003laplacian} are used to reduce dimensionality within each respective domain. Subsequently, known correspondences are used to perform Procrustes analysis~\cite{gower2004procrustes}, aligning the embeddings by minimizing differences between the corresponding points. MAPA’s results were shown to outperform SSMA in terms of mapping probabilities typically. The MAPA approach was extended in~\cite{shen2017manifold-matching} by constructing ranked distances and determining shortest paths within each domain. In this extended approach, MDS is applied to these ranked distances as a precomputed metric. Alignment is achieved via Procrustes matching of the resulting embeddings. Evaluations on the Swiss Roll dataset demonstrated high matching ratios, but no comparisons with other alignment methods were performed.

Whereas most of the preceding approaches used Laplacian Eigenmaps as the primary method for dimensionality reduction, the work of~\cite{lafon2006datafusion} employed Diffusion Maps~\cite{coifman2006dm} as a framework for MA. They introduced similarity normalization in their graph structure to ensure invariance to data density and then embedded each domain separately into a lower-dimensional space to capture intrinsic geometry. An affine transformation is learned via the Iterative Closest Point algorithm~\cite{fitzgibbon2003icp} and geometric hashing~\cite{wofson1997geom-hashing} to align the embeddings.

In~\cite{amodio2018magan}, the authors introduced a semi-supervised MA method using a pair of generative adversarial networks, called the Manifold Alignment Generative Adversarial Network (MAGAN). The method employs three loss functions to guide the alignment: (1) reconstructive loss, which ensures consistency when mapping back and forth between domains; (2) discriminator loss, which distinguishes between real and generated points; and (3) correspondence loss, which penalizes deviations from known correspondences. Although originally applied to co-domains with shared features but no explicit correspondences, we adapted MAGAN’s correspondence loss function for cases with known correspondences.

Diffusion Transport Alignment (DTA) was introduced in~\cite{duque2023dta} as a method that integrates diffusion processes with regularized optimal transport~\cite{villani2009optimaltransport, courty2014domain-reg-optimal}. It constructs local similarity matrices for each domain using a kernel function on $k$-NN graphs, which are then row-normalized to form diffusion operators ($P_{\mathcal{X}}$ and $P_{\mathcal{Y}}$). Cross-domain transition matrices, derived from known correspondences, facilitate the computation of inter-domain distances. The alignment problem is then solved using regularized optimal transport, with DTA outperforming MAGAN, MAPA, and SSMA in the authors' empirical evaluations.

More recently, two graph-based methods, MASH (Manifold Alignment via Stochastic Hopping) and SPUD (Shortest Paths on the Union of Domains), were introduced in~\cite{rhodes2024gdi}. Both leverage graph structures in manifold learning to preserve local and learn global data structures while integrating inter-domain correspondences. The methods share the same initialization strategy, but differ in how they learn cross-domain information. They begin by building weighted graphs for each domain using an $\alpha$-decaying kernel~\cite{moon2019phate} to capture local geometries. Known correspondences, referred to as anchor points, create inter-domain edges that bridge the two graphs. A combined graph is then constructed, where connectivity between domains is established via these anchors.

SPUD estimates geodesic distances via shortest paths in this graph, connecting points by traversing intra-domain paths and crossing domains through the anchors. These distances are embedded into a low-dimensional space using MDS or other embedding methods. While SPUD performs well with low noise and well-defined anchors, its reliance on shortest paths makes it vulnerable to errors in noisy or sparse graph structures.

In contrast, MASH employs a diffusion-based approach that is more robust to noise and sparsity. MASH normalizes combined similarity graphs to create a cross-domain diffusion operator that captures inter-domain transition probabilities, enabling random walks within and between the domains, using the anchor points to connect the domains. An iterative diffusion process over multiple steps produces a multiscale representation encoding both local and global structures. The resulting diffusion distances, derived from the most probable paths, filter out connections induced by noise and are embedded into a shared manifold using MDS.

Of these approaches, only MAGAN provides a natural out-of-sample extension mechanism via its generator function. Other methods, such as DTA and MASH, provide an operator (optimal transport or diffusion) that allows for a cross-domain mapping of existing points via barycentric projection, but still cannot extend to out-of-sample points.

\subsection{Cross-Domain Mapping}\label{subsec:cross_domain_mapping}

The primary result of most of the described alignment methods is an embedding in a shared space, while some, such as like MAGAN and DTA, offer mechanisms for direct cross-domain mapping. MAGAN utilizes trained generators to project data points between domains. The generators provide a direct mapping function between domains. DTA’s coupling matrix facilitates barycentric projection~\cite{duque2023dta}. This matrix effectively shuffles the indices to provide a mapping to the most similar points of the other domain.

The cross-domain diffusion operator in MASH captures instance-level attributions, indicating how points in one domain influence the generation of an embedded point in another domain. By normalizing the cross-domain entries of the $t$-step diffusion operator, a weighted average of these entries can be computed to project a representation across domains.

Our general method provides a direct mechanism for cross-domain mappings from a pre-aligned embedding space to one's space of choice via one of the decoder functions. This works similarly to MAGAN, but has the flexibility of learning an aligned space consistent with the user's choice of underlying alignment method. We compare the mapping functions to MAGAN, DTA, and MASH in Figure~\ref{fig:cross-domain-map-example}.

\begin{figure}[!htb]
    \centering
    \includegraphics[width=\linewidth]{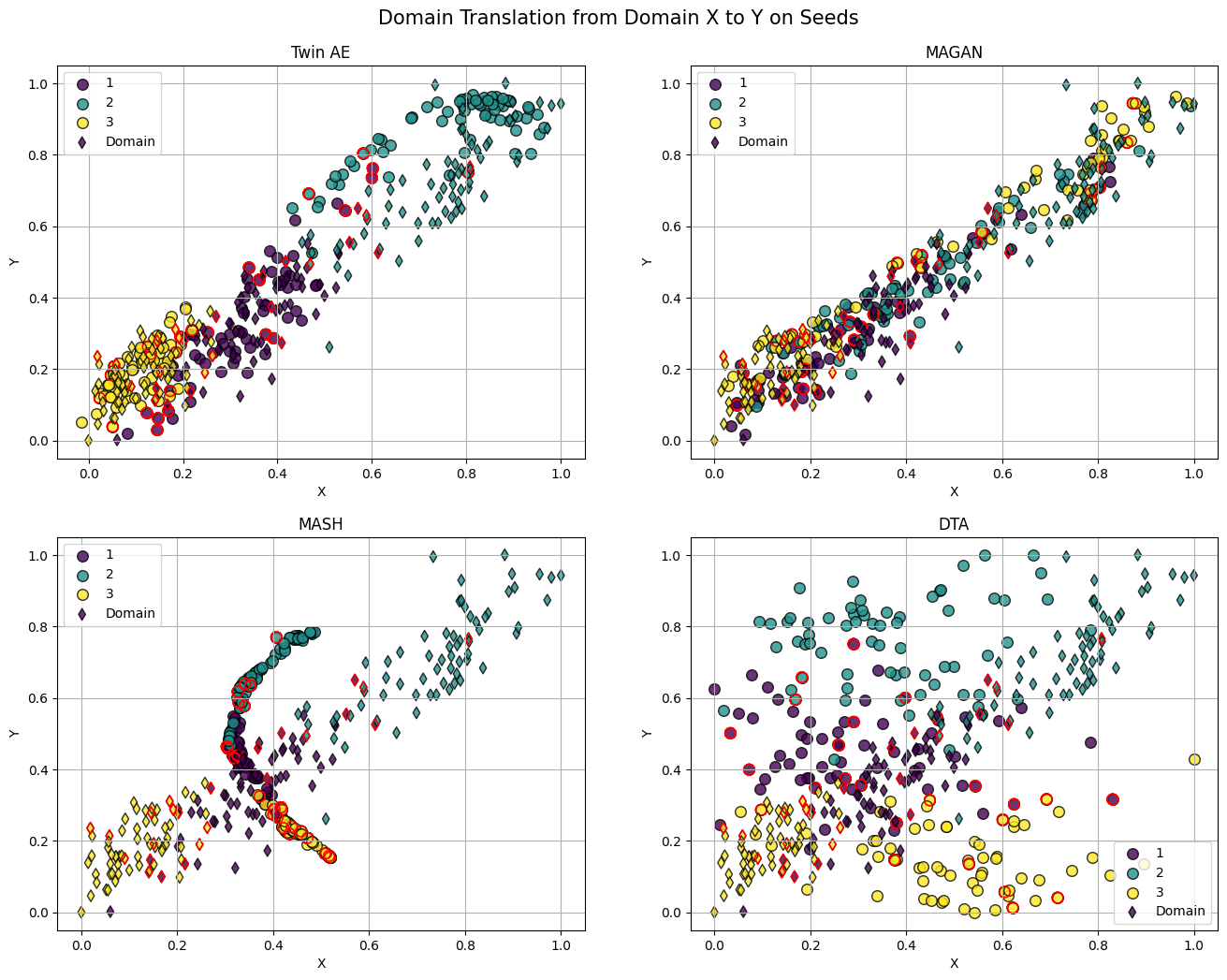}
    \caption{This figure compares how each method maps points from Domain X to Domain Y. Domain X consists of half of the features from the Seeds dataset, while the other half constitute Domain Y. Points outlined in red indicate anchors used in the alignment process. Different fill colors represent distinct class labels. Circles denote the predicted points, which originate from Domain X and are mapped to Domain Y, while diamonds represent the true data distribution in Domain Y. We observe that the twin AE  effectively translates points while maintaining their geodesic regions. Additionally, twin AE, MASH, and DTA preserve class boundaries well, whereas MAGAN’s realignment results in a more jumbled structure.}
    \label{fig:cross-domain-map-example}
\end{figure}

%%%%%%%%%%%%%%%%%%%%%%%%%%%%%%%%%%%%%%%%%%%%%%%%%%%%%
\subsection{Manifold Extension}\label{subsec:manifold_learning_extensions}
%%%%%%%%%%%%%%%%%%%%%%%%%%%%%%%%%%%%%%%%%%%%%%%%%%%%%

The inability of MA methods to extend to out-of-sample points presents an obvious weakness for real-world applications. Kernel methods for extension, such as the Nyström extension~\cite{bengio2003out}, compute similarities between new and existing points and estimate new embedded points as a function of the embedding eigenvectors. Landmark-based approaches~\cite{desilva2004sparse-mds, silva2005landmark-lasso, moon2019phate, peng2024scalablemanifoldlearninguniform} produce embeddings based on proximity to the nearest landmark. This latter approach is often marketed for computational speedup rather than usage as an out-of-sample extension~\cite{moon2019phate}.

Kernel methods have been effectively integrated into certain manifold learning techniques, such as Locally Linear Embedding (LLE) and Isomap~\cite{bengio2003out}, serving to enhance eigenvector-based approaches. However, their applicability remains largely restricted to these methods and does not extend well to diffusion-based techniques or MA frameworks aimed at connecting additional data modalities~\cite{Gigante2019CompressedD}. While kernel methods offer notable computational benefits, their ability to capture complex, nonlinear relationships is constrained by their reliance on predefined kernel functions, which may struggle to adapt to the intrinsic geometry of highly nonlinear manifolds~\cite{aizenbud2015pca}. Moreover, kernel methods often suffer from scalability issues when applied to large datasets, as the need to compute and store pairwise similarities results in significant memory and computational overhead. These drawbacks hinder their effectiveness in tasks that demand flexible, data-driven representations of manifold structures.

As an alternative approach, neural networks are used for out-of-sample extension by learning a parametric function. For example, parametric variants of $t$-SNE~\cite{van2009parametric-tsne} and UMAP~\cite{sainburg2021parametricumap} enlist feedforward neural networks to run the embedding optimization process. Another similar extension for inductive $t$-SNE trains a neural network regressor to directly predict the locations of points in the embedded space~\cite{roman-rangle2019inductive-tsne}.

Direct regression onto the embedding space often leads to solutions that overfit the observed data, resulting in poor generalization to unseen samples~\cite{arpit2017closer}. This issue arises because the network may model nuances (noise) in the training data rather than learning general, transferable representations. Moreover, using only point embeddings as responses in a loss function may not provide sufficient structural constraints, leading to degenerate mappings that fail to preserve the geometric properties intrinsic to the original data space.

To mitigate these challenges, multitask learning has proven to be an effective form of implicit regularization, helping guide the learning process toward accurate, structured, and generalizable solutions~\cite{Le2018supervised_autoencoders, zhang2018multi-task-learning}. In particular, AE-based approaches---such as geometry-regularized AEs~\cite{duque2020grae}---have demonstrated the advantages of simultaneously predicting embeddings while reconstructing the input data. This dual-task framework improves generalization and ensures that the learned representations retain essential geometric and topological properties, facilitating smooth out-of-sample extensions and cross-domain mapping.

We thus present a guided representation learning method for manifold alignment that extends to out-of-sample points through the use of AEs. Our approach employs a twin set of regularized AEs to jointly preserve the geometric structure of learned embeddings (via the encoder) and enforce a structured mapping between modalities (via the decoder). The encoder is trained to follow a pre-aligned embedding created by an alignment method of choice, while the decoder forces the learned representation to retain sufficient information to reconstruct the original data from the embedding space. This multitask formulation improves cross-domain generalization and mitigates overfitting, leading to a more accurate alignment of the data distributions. 

\section{Additional Results}\label{sec:add_results}

To quantify the comparative cross-domain mapping of MAGAN, MASH, DTA, and our twin AE model, we calculate the MSE between the mapped and original corresponding points. Regardless of the underlying MA model used to train the twin AE networks, the cross-domain mapping using the AEs after decoder swapping outperforms the projections or mappings by MASH, DTA, and MAGAN, when applied to a test set. The aggregate results---across all splits and the datasets described in Tables~\ref{tab:class_datasets} and \ref{tab:reg_data}---are shown in Figure~\ref{fig:cross-domain-loss}

\begin{figure}[!htb]
    \centering
    \includegraphics[width=0.7\linewidth]{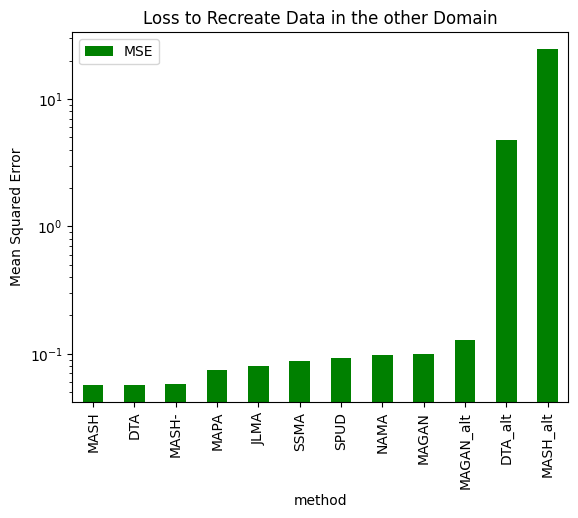}
    \caption{MSE of cross-domain mapped points (log scale) for test data. All comparisons are conducted on test points that were never part of the embedding space. MAGAN\_ALT extends points from one domain to another using the native neural networks of MAGAN. DTA\_ALT applies barycentric projection, while MASH\_ALT follows a similar approach but within the projection space learned by MASH. All other results correspond to the underlying MA models used to regularize the twin AEs. The AE approach performs better than the competing methods regardless of underlying MA model regularization, though the mapping is best when using MASH (and variants) or DTA.}
    \label{fig:cross-domain-loss}
\end{figure}

\section{Datasets}

We validate our models using publicly available datasets from the UCI repository~\cite{dua2019uci}. Tables~\ref{tab:class_datasets} and \ref{tab:reg_data} summarize the datasets employed in our experiments, listing the number of observations ($n$) and features ($d$) for each one. The collection spans a diverse range of domains and complexities, including medical data (e.g., breast cancer and diabetes), activity-related data (e.g., chess and car evaluation), and data from physical simulations (e.g., waveform analysis).

\begin{table}[!htb]
    \centering
    \scriptsize
    \caption{Classification Datasets. All datasets are from the UCI ML repository~\cite{dua2019uci} unless otherwise cited. All missing values were removed before use.}
    \begin{tabular}{llrr}
    \toprule
        Dataset & $n$ & $d$ \\
    \midrule
        Artificial Tree~\cite{moon2019phate} & 1440 & 60 \\
        Balance Scale  & 645 & 4 \\
        Breast Cancer  & 699 & 16 \\
        Car  & 1728 & 6 \\
        Chess & 3196 & 36 \\
        CRX & 664 & 14 \\
        Diabetes  & 678 & 8 \\
        Flare1 & 323 & 10 \\
        Glass  & 214 & 10 \\
        Heart Disease  & 303 & 13 \\
        Heart Failure & 299 & 10 \\
        Hepatitis & 138 & 15 \\
        Hill Valley  & 606 & 101 \\
        Ionosphere  & 351 & 34 \\
        Iris  & 150 & 4 \\
        Optical Digits  & 3823 & 64 \\
        Parkinson's  & 197 & 23 \\
        Seeds   & 210 & 7 \\
        Tic-Tac-Toe  & 958 & 9 \\
        Titanic & 712 & 7\\
        Waveform  & 5000 & 21 \\
        Wine  & 178 & 13 \\
    \bottomrule
    \end{tabular}
    \label{tab:class_datasets}
\end{table}

\begin{table}[!htb]
    \centering
    \scriptsize
    \caption{These datasets use a continuous response which was evaluated using our twin AE MA model. }
    \label{tab:reg_data}
    \begin{tabular}{llrr}
        \toprule
        Dataset & $n$ & $d$ \\
        \midrule
        Airfoil Self-Noise & 1503 & 5 \\
        Air Quality & 9357 & 12 \\
        Automobile & 159 & 18 \\
        Auto MPG & 392 & 7 \\
        Beijing PM25 & 41757 & 11 \\
        Community Crime & 1994 & 100 \\
        Computer Hardware & 209 & 6 \\
        Concrete Compressive Strength & 1030 & 8 \\
        Concrete Slump Test & 103 & 9 \\
        Cycle Power Plant & 9568 & 4 \\
        Energy Efficiency & 768 & 8 \\
        Facebook Metrics & 495 & 17 \\
        Five Cities PM25 & 21436 & 9 \\
        Hydrodynamics & 308 & 6 \\
        Istanbul Stock & 536 & 8 \\
        Naval Propulsion Plants & 11934 & 16 \\
        Optical Network & 630 & 9 \\
        Parkinsons & 5875 & 21 \\
        Protein & 45730 & 9 \\
        SML 2010 & 4137 & 21 \\
        \bottomrule
    \end{tabular}
\end{table}

\section{ADNI Variables and Results}\label{sec:adni_vars}

Data used in the preparation of this article were obtained from the Alzheimer's Disease Neuroimaging Initiative (ADNI) database (\href{adni.loni.usc.edu}{adni.loni.usc.edu}). The ADNI was launched in 2004 as a public-private partnership, led by Principal Investigator Michael W. Weiner, MD. The original goal of ADNI was to test whether serial magnetic resonance imaging (MRI), positron emission tomography (PET), other biological markers, and clinical and neuropsychological assessment can be combined to measure the progression of mild cognitive impairment (MCI) and early Alzheimer's disease (AD). The current goals include validating biomarkers for clinical trials, improving the generalizability of ADNI data by increasing diversity in the participant cohort, and to provide data concerning the diagnosis and progression of Alzheimer’s disease to the scientific community. For up-to-date information, see \href{adni.loni.usc.edu}{adni.loni.usc.edu}.

Data collection and sharing for the Alzheimer's Disease Neuroimaging Initiative (ADNI) is funded by the National Institute on Aging (National Institutes of Health Grant U19AG024904). The grantee organization is the Northern California Institute for Research and Education. In the past, ADNI has also received funding from the National Institute of Biomedical Imaging and Bioengineering, the Canadian Institutes of Health Research, and private sector contributions through the Foundation for the National Institutes of Health (FNIH) including generous contributions from the following: AbbVie, Alzheimer’s Association; Alzheimer’s Drug Discovery Foundation; Araclon Biotech; BioClinica, Inc.; Biogen; BristolMyers Squibb Company; CereSpir, Inc.; Cogstate; Eisai Inc.; Elan Pharmaceuticals, Inc.; Eli Lilly and Company; EuroImmun; F. Hoffmann-La Roche Ltd and its affiliated company Genentech, Inc.; Fujirebio; GE Healthcare; IXICO Ltd.; Janssen Alzheimer Immunotherapy Research \& Development, LLC.; Johnson \& Johnson Pharmaceutical Research \& Development LLC.; Lumosity; Lundbeck; Merck \& Co., Inc.; Meso Scale Diagnostics, LLC.; NeuroRx Research; Neurotrack Technologies; Novartis Pharmaceuticals Corporation; Pfizer Inc.; Piramal Imaging; Servier; Takeda Pharmaceutical Company; and Transition Therapeutics.

\begin{table}[h]
    \centering
    \scriptsize
    \caption{Variables included in the FAQ domain. All codings are as follows:
    0 = Normal; 1 = Never did, but could do now; 2 = Never did, would have difficulty now; 
    3 = Has difficulty, but does by self; 4 = Requires assistance; 5 = Dependent. 
    The root mean squared error (RMSE) of the cross-domain mapping is given. Since the FAQ 
    items are scored on a 0--5 scale, RMSE values around 1 represent roughly 20\% of the 
    total range, indicating moderate prediction error. Smaller RMSE values (e.g., $\approx$0.8, 
    or $\approx16$\% of the scale) indicate relatively more accurate prediction, whereas values above 
    1.2 ($\approx$24\% of the scale) reflect greater difficulty in approximating those items.}
    \begin{tabular}{|p{1.5cm}|p{5.8cm}|c|}
        \hline 
        Variable & Question Target & RMSE \\ \hline 
        FAQFINAN & Writing checks, paying bills, or balancing checkbook. & 1.104 \\ \hline 
        FAQFORM  & Assembling tax records, business affairs, or other papers. & 1.210 \\ \hline 
        FAQSHOP  & Shopping alone for clothes, household necessities, or groceries. & 0.913 \\ \hline 
        FAQGAME  & Playing a game of skill (e.g., bridge, chess) or hobby. & 0.913 \\ \hline 
        FAQBEVG  & Heating water, making coffee, turning off the stove. & 0.829 \\ \hline 
        FAQMEAL  & Preparing a balanced meal. & 1.036 \\ \hline 
        FAQEVENT & Keeping track of current events. & 0.952 \\ \hline 
        FAQTV    & Understanding a TV program, book, or magazine. & 0.967 \\ \hline 
        FAQREM   & Remembering appointments, occasions, holidays, medications. & 1.391 \\ \hline 
        FAQTRAVL & Traveling, driving, or arranging public transport. & 1.083 \\ \hline
    \end{tabular}
    \label{tab:FAQ_Variables}
\end{table}

\begin{table}[h]
    \centering
    \scriptsize
    \caption{Variables included in the ADAS-Cog 13 Domain. All scores range from zero to the maximum score described. The root mean squared error (RMSE) of the cross-domain mapping using the FAQ domain onto the ADAS-Cog values is given. Lower values correspond to smaller errors relative to the task’s scale. For example, an RMSE of 0.45 on a 5-point task ($\approx$9\% of the range) indicates relatively accurate prediction, while an RMSE of 2.35 on a 12-point task ($\approx$20\% of the range) reflects greater difficulty in approximating that score. In general, RMSE values below 1 correspond to relatively small errors for their ranges, whereas values above 2 correspond to more substantial deviations.}
    \begin{tabular}{|c|p{4.5cm}|c|c|}
        \hline
        Var & Task & Max & RMSE \\ \hline
        Q1SCORE  & Word Recall         & 10 & 1.459 \\ \hline
        Q2SCORE  & Commands            & 5  & 0.449 \\ \hline
        Q3SCORE  & Construction        & 5  & 0.675 \\ \hline
        Q4SCORE  & Delayed Word Recall & 10 & 2.344 \\ \hline
        Q5SCORE  & Naming              & 5  & 0.447 \\ \hline
        Q6SCORE  & Ideational Praxis   & 5  & 0.422 \\ \hline
        Q7SCORE  & Orientation         & 8  & 0.816 \\ \hline
        Q8SCORE  & Word Recognition    & 12 & 2.347 \\ \hline
        Q9SCORE  & Recall Instructions & 5  & 0.523 \\ \hline
        Q10SCORE & Spoken Language     & 5  & 0.325 \\ \hline
        Q11SCORE & Word Finding        & 5  & 0.610 \\ \hline
        Q12SCORE & Comprehension       & 5  & 0.416 \\ \hline
        Q13SCORE & Number Cancellation & 5  & 0.851 \\ \hline
    \end{tabular}
    \label{tab:ADAS_Variables}
\end{table}

%%%%%%%%%%%%%%%%%%%%%%%%%%%%%%%%%%%%%%%%%%%%%%%%%%%%%
% \section*{References}
%%%%%%%%%%%%%%%%%%%%%%%%%%%%%%%%%%%%%%%%%%%%%%%%%%%%%

}

\end{document}